\documentclass[final,5p,times,twocolumn]{elsarticle}
\usepackage{cite}
\usepackage{amsmath,amssymb,amsfonts}
\usepackage{algcompatible}
\usepackage{algorithm}

\usepackage{bm}
\usepackage{color}

\newtheorem{theorem}{Theorem}
\newtheorem{lemma}{Lemma}
\newdefinition{definition}{Definition}
\newtheorem{corollary}{Corollary}
\newtheorem{proposition}{Proposition}
\newdefinition{remark}{Remark}
\newdefinition{assumption}{Assumption}

\newproof{proof}{Proof}

\DeclareMathOperator{\E}{\mathbb{E}}

\DeclareMathOperator{\card}{card}

\newcommand{\qedproof}{\hfill\blacksquare}
\newcommand{\xx}{\mathbf{x}}
\newcommand{\uu}{\mathbf{u}}
\newcommand{\RR}{\mathbb{R}}
\newcommand{\NN}{\mathbb{N}}
\newcommand{\XX}{\mathcal{X}}
\newcommand{\UU}{\mathcal{U}}
\renewcommand{\AA}{\mathcal{A}}
\renewcommand{\SS}{\mathcal{S}}

\journal{}

\begin{document}

\begin{frontmatter}

\title{Approximated Multi-Agent Fitted Q Iteration}

\author[label1]{Antoine Lesage-Landry}
\ead{antoine.lesage-landry@polymtl.ca}
\author[label2]{Duncan S. Callaway}
\ead{dcal@berkeley.edu}

\address[label1]{Department of Electrical Engineering, Polytechnique Montréal, Mila \& GERAD, 2500 de Polytechnique Road, Montréal,H3T 1J4, Québec, Canada}

\address[label2]{Energy and Resources Group, University of California, Berkeley, 310 Barrows Hall, Berkeley, 94720, California, U.S.A.}

\begin{abstract}
We formulate an efficient approximation for multi-agent batch reinforcement learning, the approximated multi-agent fitted Q iteration (\texttt{AMAFQI}). We present a detailed derivation of our approach. We propose an iterative policy search and show that it yields a greedy policy with respect to multiple approximations of the centralized, learned Q-function. In each iteration and policy evaluation, \texttt{AMAFQI} requires a number of computations that scales linearly with the number of agents whereas the analogous number of computations increase exponentially for the fitted Q iteration (\texttt{FQI}), a commonly used approaches in batch reinforcement learning. This property of \texttt{AMAFQI} is fundamental for the design of a tractable multi-agent approach. We evaluate the performance of \texttt{AMAFQI} and compare it to \texttt{FQI} in numerical simulations. The simulations illustrate the significant computation time reduction when using \texttt{AMAFQI} instead of \texttt{FQI} in multi-agent problems and corroborate the similar performance of both approaches.

\end{abstract}


\end{frontmatter}

\section{Introduction}
\label{sec:intro}

Reinforcement learning is a framework which considers stochastic, sequential decision-making problems with unknown dynamics~\citep{sutton2018reinforcement}. These problems are modelled as Markov decision processes (MDPs). In each decision round of an MDP, a decision maker observes the current state of the system and must provide a decision or equivalently, a control. A scalar reward is subsequently revealed, and the current state shifts to a new state according to a transition function defined by the dynamics of the problem. In reinforcement learning, the transition function is unknown. Only the reward, the initial and resulting states, and the control are used to improve future controls. Batch reinforcement learning~\citep{lange2012batch,ormoneit2002kernelbased,ernst2005tree} is a subfield of reinforcement learning in which information about the system in the form of a set of historical transitions is known a priori to the decision maker. This is in contrast to typical reinforcement learning algorithms, e.g., the $Q$-learning algorithm~\citep{watkins1992q}, in which information is gathered in an online fashion. Batch reinforcement learning improves over its online counterpart (i) by reusing the gathered information multiple times (experience replay~\citep{lin1992self}) to increase the approach's convergence speed, (ii) by fitting an approximated function (e.g., $Q$ or value functions) in between updates to mitigate instabilities, and (iii) by averaging similar transitions from the batch information to better estimate the MDP's stochastic model~\citep{lange2012batch}. In batch reinforcement learning, a prevalent approach~\citep{lange2012batch} is the fitted $Q$ iteration (\texttt{FQI})~\citep{ernst2005tree}.

In multi-agent reinforcement learning, agents make sequential decisions to maximize their joint or individual rewards~\citep{bu2008comprehensive,zhang2018fully}. The agents can be fully cooperative, i.e., maximizing a joint reward function, fully competitive, i.e., the agents' objectives are opposed, or a combination of both~\citep{bu2008comprehensive,zhang2018fully}. The main challenge when considering the multi-agent reinforcement learning problem comes from the cardinality of the joint control set as it increases exponentially with the number of agents. This adds to the difficulty that the curse of dimensionality already poses to (approximate) dynamic programming-based methods~\citep{shoham2003multi,bu2008comprehensive,zhang2018fully}. The design of an approach that relies only on local control sets is, therefore, highly desirable to enable the implementation of batch reinforcement learning methods in real-world multi-agent systems, e.g., electric power systems~\citep{callaway2010achieving}. For example, the approach we present in this work could extend current methods for demand response or distributed energy resource management like~\citep{vandael2015reinforcement,ruelens2016residential,mbuwir2017battery} to multi-agent implementations and increase the benefits for the electric grid without significantly impacting the computational cost of the approach. Other applications for multi-agent reinforcement learning include the control of a robot team~\citep{stone2000multiagent} or of an autonomous vehicle fleet~\citep{pham2018cooperative}, autonomous driving~\citep{shalev2016safe}, and stock trading~\citep{lee2007multiagent}. In this work, we consider the batch reinforcement learning framework and design the approximated multi-agent fitted $Q$ iteration (\texttt{AMAFQI}), an efficient approximation of \texttt{FQI}~\citep{ernst2005tree} tailored to fully cooperative, multi-agent problems.

\subsubsection*{Related work}
Multi-agent reinforcement learning has been studied by many authors and the main recent advancements to this body of work are reviewed in~\citep{bu2008comprehensive,zhang2019multi,oroojlooyjadid2019review,hernandez2019survey}. Multi-agent extensions to the $Q$-learning algorithm~\citep{watkins1992q} are reviewed in~\citep{bu2008comprehensive}. Reference~\citep{zhang2019multi} focuses on theory-backed approaches. An overview of multi-agent deep reinforcement learning is presented in~\citep{oroojlooyjadid2019review,hernandez2019survey}. 
In our work, we are interested in multi-agent extensions of batch reinforcement learning~\citep{lange2012batch}, and more specifically, of the kernel-based~\citep{ormoneit2002kernelbased} \texttt{FQI}~\citep{ernst2005tree} framework. Multi-agent problems have also been studied under other reinforcement learning frameworks, e.g., classical $Q$-learning~\citep{boutilier1996planning,lauer2000algorithm} or actor-critic approaches~\citep{gupta2017cooperative,zhang2018fully}. We review the literature relevant to multi-agent \texttt{FQI} next.

To the best of the authors' knowledge, the only extension of \texttt{FQI} to the multi-agent setting are presented in~\citep{gabel2008evaluation,gabel2008reinforcement,zhang2021finite}. References~\citep{gabel2008evaluation,gabel2008reinforcement} only consider deterministic problems. The extension relies on the neural fitted $Q$ (\texttt{NFQ}) algorithm~\citep{riedmiller2005neural}. The \texttt{NFQ} is a modified \texttt{FQI} approach that uses a neural network instead of a regression tree as the fitting method used to generalize the $Q$-value to all state-control pairs (see Section~\ref{sec:fqi}). Similarly to our approach, their work is based on the ideas of~\citep{lauer2000algorithm} in which an efficient multi-agent $Q$-learning algorithm~\citep{watkins1992q}, the Distributed $Q$-learning algorithm, for online, deterministic settings is presented, to obtain an approach that does not require computations over the joint control set. The work of~\citep{gabel2008evaluation,gabel2008reinforcement} differs from ours because it uses an opportunistic approach enabled by the deterministic setting.
Furthermore,~\citep{gabel2008evaluation,gabel2008reinforcement} only provide an empirical analysis of their algorithm because the properties of the neural network are hard to analyze. In our work, we (i) consider general stochastic problems, (ii) present a detailed derivation for \texttt{AMAFQI}, and (iii) provide a convergence analysis of the approximated local $Q$-functions used by our approach. Moreover, we characterize the performance of the greedy policy for \texttt{AMAFQI}. We omit the fitting or regression error from the performance analysis of \texttt{AMAFQI} similarly to \texttt{FQI}'s~\citep{ernst2005tree}. This will be investigated in future work. Lastly,~\citep{zhang2021finite} proposes a general, fully decentralized, multi-agent fitted $Q$ algorithm that accounts for competitive agents and where any function approximator can be used to approximated the local $Q$-function. The authors further derive a finite-sample performance guarantee for their approach. However,~\citep{zhang2021finite}'s algorithm requires optimizing local $Q$-function over the joint control space which grows exponentially with the number of agents. Our main contribution is to provide an approach which uses only local control spaces.

Our specific contributions are:
\begin{itemize}
    \item We formulate the approximated multi-agent fitted $Q$ iteration (\texttt{AMAFQI}). \texttt{AMAFQI} is an efficient approximation of the \texttt{FQI} algorithm for multi-agent settings. In each iteration, \texttt{AMAFQI}'s computation scales linearly in the number of agents instead of exponentially as in \texttt{FQI}.
    \item We propose a policy search for \texttt{AMAFQI} and show that it is a greedy policy with respect to the approximation of the centralized, learned $Q$-functions from each agent. 
    \item We derive a very efficient extension of \texttt{AMAFQI}, \texttt{AMAFQI-L}, that further reduces the computation requirement of the approach.
    \item We show the convergence of the local $Q$-function approximations computed by \texttt{AMAFQI} to unique and finite functions.
    \item We numerically evaluate the performance of \texttt{AMAFQI}. We show the similar performance and significant decrease in computation times when \texttt{AMAFQI} and \texttt{AMAFQI-L} are used instead of \texttt{FQI}.
\end{itemize}

\section{Preliminaries}
\label{sec:prelim}

We consider an MDP ($\XX, \UU, f, r$) where multiple agents must implement a control to maximize their expected joint cumulative reward. Let $m \in \NN$ be the number of agents. We assume $m > 1$.
Let $\XX \subseteq \RR^{n \times m}$, $\UU \subseteq \RR^{p \times m}$, and $\mathcal{W} \subseteq \RR^{s \times m}$ where $n,p,s \in \NN$ be the joint state,  control, and disturbance space, respectively. Let $\xx \in \XX$ be a joint state, $\uu \in \UU$ be a joint control, and $\mathbf{w} \in \mathcal{W}$ be a random disturbance. Let $f: \XX \times \UU \times \mathcal{W} \mapsto \XX$ express the state transition function of the problem. The function $f$ maps an initial state, a control and a disturbance to a resulting state. Lastly, let $r: \XX \times \UU \times \mathcal{W} \mapsto \RR$ be the function that returns the reward associated with an initial state, control, final state, and disturbance tuple. We make the following assumption regarding the reward function.
\begin{assumption}
The reward function $r$ is bounded below and above such that $0 \leq r(\xx, \uu, \mathbf{w}) \leq R < +\infty$ for all $(\xx,\uu,\mathbf{w}) \in \XX \times \UU \times \mathcal{W}$. \label{ass:reward}
\end{assumption}
The assumption on the upper bound of the reward function is a standard assumption for MDPs in reinforcement learning~\citep{ernst2005tree}. The lower bound assumption is mild because if not met, a constant can be added to the reward function so that it is non-negative. This translation does not change the optima~\citep{lauer2000algorithm}.

To easily differentiate local and joint controls, we define local control variables and spaces. We let $\AA^j \subset \RR^p$ be the local control space of agent $j$ where $\UU = \times_{j=1}^m \AA^j$. We denote a local control by $a \in \AA^j$ and add the superscript $j$ to refer to the $j^{\text{th}}$ agent if needed. 

Formally, the $m$ agents want to cooperatively solve the following problem:
\begin{equation}
\max_{\left\{\uu_T \in \UU\right\}_{T=1}^{+\infty}} \E \left[ \sum_{T=1}^{+\infty} \beta^{T} r(\xx_T, \uu_T, \mathbf{w}_T) \right]
\label{eq_the_problem}
\end{equation}
where $\beta \in [0 , 1)$ is the discount factor. The variables $\uu_T$ and $\xx_T$ represent the joint control and state at the decision round $T$, respectively. The random disturbance at $T$ is represented by $\mathbf{w}_T$. Successive states are obtained from $\xx_{T+1} = f(\xx_{T}, \uu_{T}, \mathbf{w}_{T})$, where $\mathbf{w}_T \in \mathcal{W}$. The expectation in~\eqref{eq_the_problem} is taken with respect to the probability of $\mathbf{w}_T$ given the state and control at round $T$. 

We consider the batch reinforcement learning framework~\citep{lange2012batch,ormoneit2002kernelbased,ernst2005tree}. In this setting, $f$ is unknown and only examples of past transitions can be used to solve~\eqref{eq_the_problem}. The decision makers or agents have access to batch data representing historical transitions~\citep{ernst2005tree}. The batch data is used to first compute an approximation of the $Q$-function and, second, to evaluate a policy. 
Let $L \in \NN$ be the number of available samples in the batch data. The batch data set $\SS_L$ is defined as:
\begin{align*}
\SS_L = \left\{ \left(\xx^l , \uu^l, \xx_{+}^l, r^l \right)\in \XX \times \UU \times \XX \times \RR_+,l = 1, 2 ,\ldots, L \right\},
\end{align*}
where $\xx_+^l$ refers to the state observed after control $\uu^l$ was implemented in state $\xx^l$. These samples do not need to be generated from continuous experiments. Specifically, we focus on regression tree-based \texttt{FQI} approaches~\citep{ernst2005tree}. \texttt{FQI} is introduced in detail in the next subsection.

\subsection{Fitted Q iteration}
\label{sec:fqi}
We recall the motivation for \texttt{FQI} as presented in~\citep{ernst2005tree}. The state-action value or $Q$-function $Q:\XX \times \UU \mapsto \RR$ is the unique solution to the Bellman equation:
\[
Q\left(\xx , \uu \right) = \E\left[ r(\xx, \uu, \mathbf{w}) + \beta \max_{\uu' \in \UU} Q\left(f(\xx, \uu, \mathbf{w}), \uu' \right) \right],
\]
where $\beta \in [0,1)$. The expectation is taken with respect to the probability of $\mathbf{w}$ given the state $\xx$ and control $\uu$. By the contraction mapping theorem~\citep{luenberger1997optimization}, the $Q$-function can be obtained by successively solving
\begin{equation}
Q_N\left(\xx , \uu \right) = \E\left[ r(\xx, \uu, \mathbf{w}) + \beta \max_{\uu' \in \UU} Q_{N-1}\left(f(\xx, \uu, \mathbf{w}), \uu' \right) \right], \label{eq:Q_stoch}
\end{equation}
for all $N \geq 1$ with the boundary condition $Q_0\left(\xx , \uu \right) = 0$ for all $(\xx, \uu) \in \XX \times \UU$. In the deterministic case,~\eqref{eq:Q_stoch} can be expressed as:
\begin{equation*}
Q_N\left(\xx , \uu \right) = r(\xx, \uu) + \beta \max_{\uu' \in \UU} Q_{N-1}\left(\delta(\xx, \uu), \uu' \right), 
\end{equation*}
where $\delta:\XX \times \UU \mapsto \XX$ is the deterministic function that returns the resulting state given a pair state-control.
Given $\mathcal{S}_L$ and supposing $Q_{N-1}$ is available, then for all data points $l=1,2,\ldots, L$, we can compute
\begin{equation}
Q_N\left(\xx^l , \uu^l \right) = r^l + \beta \max_{\uu' \in \UU} Q_{N-1}\left(\xx^l_{+}, \uu' \right), \label{eq:Q_pairs}
\end{equation}
because $r(\xx^l, \uu^l) = r^l$ and $\delta(\xx^l,\uu^l) = \xx_{+}^l$. The \texttt{FQI} then works in the following way. Pairs of $\left(\xx^l, \uu^l \right)$ and their respective $Q_N\left(\xx^l , \uu^l \right)$-value can be generated using~\eqref{eq:Q_pairs} for all $l$ in the batch data. Then, an approximation $\hat{Q}_N^\text{FQI}(\xx, \uu)$ of $Q_N(\xx,\uu)$ is obtained by fitting a function over the pairs $\left(\left(\xx^l, \uu^l \right), Q_N\left(\xx^l , \uu^l \right) \right) $ for $l=1,2,\ldots,L$. This is done to estimate the state-action values for all state-control pairs based on the batch data. Using~$\hat{Q}_{N-1}^\text{FQI}$ in~\eqref{eq:Q_pairs} instead of $Q_{N-1}$, we can compute the state-action values at $N$, fit a function again based on the new pairs and obtain $\hat{Q}_{N}^\text{FQI}$. This process is then repeated until convergence. Finally, the authors of~\citep{ernst2005tree} argue that the process described above provides an adequate approximation $\hat{Q}_N^\text{FQI}(\xx, \uu)$ for the stochastic case as well. In the stochastic case, the conditional expectation of~\eqref{eq:Q_pairs}'s right-hand side given the current state and control is required for the update. Least squares regression~\citep{ernst2005tree} or the averaging at leaf nodes of regression tree methods~\citep{lange2012batch} estimates the conditional expectation of the dependent variables given the independent variables, respectively the $\hat{Q}_N^\text{FQI}(\xx^l, \uu^{l})$ and $\left(\xx^l, \uu^l, \xx_{+}^l, r^l \right)$ in this setting. Least squares and tree regression methods hence approximate the right-hand side of~\eqref{eq:Q_pairs} in the stochastic case~\citep{lange2012batch,ernst2005tree}.

\subsection{Regression tree methods}
In this work, we use a regression tree to generalize the local $Q$-function and state-control pairs. Regression trees are chosen as the regression methods because (i) their properties allow us to establish the \texttt{AMAFQI}'s convergence (see Section~\ref{sec:convergence}) and (ii) they are computationally efficient, scalable and robust to noisy data~\citep{ernst2005tree}. We now introduce regression tree methods. Let $\mathcal{I}\subseteq \RR^{n+p}$ and $\mathcal{O} \in \RR$ be, respectively, the input and output sets of the data set $\mathcal{D}=\left\{\left( i^l, o^l \right) \in \mathcal{I} \times \mathcal{O}, l=1,2,\ldots, L \right\}$. Regression tree methods subdivide the input set into partitions of input points $i^l$ using binary splits. Each partition is then given a unique output value. In regression trees, this is typically the average of all output points $o^l$ belonging to the partition. Multiple techniques exist to generate regression trees, e.g., \texttt{KD-Tree}~\citep{bentley1975multidimensional}, \texttt{CART}~\citep{breiman1984classification}, \texttt{Totally Randomized Trees}~\citep{geurts2006extremely}, or \texttt{Extra-Trees}~\citep{geurts2006extremely}. The reader is referred to~\citep{james2013introduction} for a detailed description of regression trees. 
We now state relevant properties and assumptions which we use in the next sections.

Using a regression tree method, a function $\hat{h}:\mathcal{I} \mapsto \mathcal{O}$ fitted to the data set $\mathcal{D}$ can be expressed as~\citep{ernst2005tree}: $\hat{h}(i) = \sum_{l=1}^L \text{kernel}\left(i^l ; i\right) o^{l}$, for $i \in \mathcal{I}$. The kernels are defined by: $\text{kernel}\left(i^l ; i\right) = \frac{\mathbb{I}_{i^l \in \mathcal{P}(i)}}{\sum_{\left(\hat{\imath},\hat{o} \right) \in \mathcal{D}} \mathbb{I}_{i^l \in \mathcal{P}(\hat{\imath})}}$, where $\mathbb{I}_x$, the indicator function, returns $1$ if $x$ is true and $0$ otherwise, and $\mathcal{P}(i)$ returns the tree partition input $i$ is part of. For ensemble methods, the kernels are: $\text{kernel}\left(i^l ; i\right) = \frac{1}{e} \sum_{k=1}^{e}  \frac{\mathbb{I}_{i^l \in \mathcal{P}_k(i)}}{\sum_{\left(\hat{\imath},\hat{o} \right) \in \mathcal{D}} \mathbb{I}_{i^l \in \mathcal{P}_k(\hat{\imath})}}$, where the subscript $k$ refers to the $k^\text{th}$ regression tree of the ensemble which consists of $e$ trees.

In this work, two assumptions about the regression method we use are made. These assumptions are similar to~\citep{ernst2005tree}.

\begin{assumption}
The kernels and batch data used to fit them are the same in all iterations $N$ of \texttt{AMAFQI}.
\label{ass:same}
\end{assumption}

\begin{assumption}\label{ass:normalized}
The kernels are normalized, i.e., $\sum_{l=1}^L \text{kernel}\left(i^l ; i\right) = 1 \ \forall i \in \mathcal{I}$.
\end{assumption}
Moreover, the aforementioned definition of the kernel implies that the sum of the kernel's absolute value is also one when Assumption~\ref{ass:normalized} is satisfied because kernels are nonnegative.

As noted by~\citep{ernst2005tree}, Assumption~\ref{ass:same} is satisfied naturally by a tree method like the \texttt{KD-Tree}. If the partitions generated by the tree method are random or depend on the output, this assumption can be met by computing the partitions and thus the kernels only once, i.e., when the first \texttt{AMAFQI} iteration is performed. This is the case, for example, for \texttt{Totally Randomized Trees}~\citep{geurts2006extremely} which we use in Section~\ref{sec:num}. Regression tree approaches satisfy Assumption~\ref{ass:normalized} by construction~\citep{ormoneit2002kernel,ormoneit2002kernelbased,ernst2005tree}.

\section{Approximated Multi-agent Fitted Q iteration}
\label{sec:amafqi}
We now present our multi-agent approximation of \texttt{FQI}, \texttt{AMAFQI}. The fitting iterations and policy evaluation of \texttt{AMAFQI} only depend on the local control space of the agents and do not necessitate computations over the joint control space as would require \texttt{FQI}. This allows \texttt{AMAFQI} to be a tractable multi-agent approach for batch reinforcement learning problems because optimizing a fitted $Q$-function, e.g., in~\eqref{eq:Q_pairs}, must be done by enumeration due to the use of regression trees. The cardinality of the joint control space increases exponentially with the number of agents and the cardinality of the local control space. For \texttt{FQI}, this thus leads to a prohibitively large number of calculations when computing approximated $Q$-functions and when evaluating the policy in multi-agent settings. In the next subsections, we derive the \texttt{AMAFQI} algorithm and propose a greedy policy search for our approach. We note that the approximation error due to the used of regression tree methods is neglected in this work. This is in line with~\citep{ernst2005tree}'s analysis of \texttt{FQI}.

\subsection{Derivation}
First, recall the standard $Q$-learning~\citep{watkins1992q} update for deterministic settings~\citep{lauer2000algorithm}:
\begin{equation}
Q_N\left(\xx, \uu\right) = \begin{cases}
Q_{N-1}\left(\xx, \uu\right), \qquad \text{ if } \xx \neq \xx_N \text{ or } \uu \neq \uu_N\\
r\left(\xx , \uu  \right) + \beta \max_{\uu' \in \UU} Q_{N-1}\left(\delta(\xx, \uu), \uu' \right),\\
\phantom{Q_{N-1}\left(\xx, \uu\right),} \qquad \text{ if } \xx = \xx_N \text{ and } \uu = \uu_N,
\end{cases} \label{eq:Q}
\end{equation}
with $Q_0\left(\xx, \uu\right) = 0$ for all $\left(\xx, \uu\right) \in \XX \times \UU$. We remark that in the deterministic setting, the reward $r$ is not a function of the disturbance $\mathbf{w}$. Second, consider for all agent $j=1,2,\ldots, m$, the distributed $Q$-learning update for deterministic settings~\citep{lauer2000algorithm}:
\begin{equation}
q_N^j\left(\xx, a\right) = \begin{cases}
q_{N-1}^j\left(\xx, a\right), \quad\text{ if } \xx \neq \xx_N \text{ or } a \neq \uu_N(j)\\
\max\left\{ q_{N-1}^j\left(\xx, a\right), r\left(\xx , \uu  \right) \right.\\
\left. \qquad\qquad + \beta \max_{a' \in \AA^j} q_{N-1}^j\left(\delta(\xx, \uu), a' \right)\right\},\\
\phantom{q_{N-1}^j\left(\xx, a\right),}\quad \text{ if } \xx = \xx_N, \uu = \uu_N, \text{ and } a = \uu_N(j),
\end{cases} \label{eq:d-q}
\end{equation}
with $q_0^j\left(\xx, a\right) = 0$ for all $\left(\xx, a\right) \in \XX \times \AA$. We refer to $q_{N}^j$ as local $q$-functions. The proposition below establishes a relation between the centralized and distributed updates.
\begin{proposition}{\citep[Proposition 1]{lauer2000algorithm}}
Let $(\xx, a) \in \XX \times \AA$ and suppose that $r\left( \xx, \uu\right) \geq 0$ for all $\left( \xx, \uu\right) \in \XX \times \UU$. Then, for a deterministic, fully cooperative problem, we have
\[
q_{N}^j(\xx, a) = \max_{\substack{\uu \in \UU\\ \uu(j) = a}} Q_N \left(\xx, \uu \right),
\]
for all $j=1,2,\ldots,m$ and $N \in \NN$, where $Q_N$ and $q_{N}^j$ are computed using~\eqref{eq:Q} and~\eqref{eq:d-q}, respectively.
\label{prop:fromLauer}
\end{proposition}

Let $N \in \NN$ and $j \in \left\{1,2,\ldots, m\right\}$. Consider the sample point $\left(\xx^l , \uu^l, \xx_{+}^l, r^l \right) \in \SS_L$. For now, let's assume that the function $q_{N-1}^j\left(\xx, a \right)$ is known. We define
\begin{align*}
o^{l,j}_N &= q^j_N\left(\xx^l , \uu^l(j) \right)\\
&= \max\left\{q_{N-1}^j\left(\xx^l, \uu^l(j) \right), r^l + \beta \max_{a' \in \AA^j} q_{N-1}^j\left(\xx_{+}^l, a' \right) \right\},
\end{align*}
where $\uu^l(j)$ is the $j^\text{th}$ component of the joint control $\uu^l$, i.e., the control implemented by agent $j$. Proposition~\ref{prop:fromLauer} leads to
\[
o^{l,j}_N = q^j_N\left(\xx^l , \uu^l(j) \right) = \max_{\substack{\uu \in \UU \\ \uu(j) = a}} Q_N \left(\xx^l, \uu \right),
\]
where $Q_N$ is computed via~\eqref{eq:Q}.

We now depart from prior multi-agent reinforcement learning approaches to derive \texttt{AMAFQI}. We apply the reasoning behind \texttt{FQI}~\citep{ernst2005tree} to compute an approximation $\hat{q}^j$ of the local $q^j$-function. This is done iteratively. First, we compute the $q^j$-function values at each batch data point using~\eqref{eq:d-q}. Second, we fit the approximation function $\hat{q}^j_N\left(\xx, a \right)$ to the set $\left\{ \left(\left(\xx^l, \uu^l(j) \right), \hat{q}_{N}^j\left(\xx^l, \uu^l(j) \right) \right), l=1,2,\ldots, L\right\}$ using a regression tree method. Specifically, at iteration $N \in \NN$ and for all samples $l = 1, 2, \ldots, L$, let,
\begin{align*}
i^{l,j} &= \left(\xx^l, \uu^l(j) \right)\\
o^{l,j}_N &= \max\left\{\hat{q}_{N-1}^j\left(\xx^l, \uu^l(j) \right), r^l + \beta \max_{a' \in \AA^j} \hat{q}_{N-1}^j\left(\xx_{+}^l, a' \right) \right\},
\end{align*}
where $\hat{q}_{0}^j\left(\xx, a \right) = 0$ for all $\left(\xx, a \right) \in \XX \times \AA$. Then, we compute
\begin{align}
\hat{q}^j_N\left( \xx, a  \right) = &\text{RegressionTree}\left( \left\{\left(i^{l,j} , o^{l,j}_N \right), l =1, 2,\ldots, L\right\}; (\xx , a) \right) \label{eq:q_hat_first}
\end{align}
Equivalently, we can express~\eqref{eq:q_hat_first} as
\begin{equation}
\hat{q}^j_N\left( \xx, a  \right) = \sum_{l=1}^L \text{kernel}\left(\left(\xx^l,\uu^l(j)\right);(\xx,a)\right) o^{l,j}_N, \label{eq:q_hat}
\end{equation}
for all $j=1,2,\ldots,m$. The \texttt{FQI}-based approach is used to generalize the information obtained from the batch data to all state-control pairs~\citep{ernst2005tree}. The regression step estimates values of the local $\hat{q}^j$-function and thus approximates the maximum of the $Q$-function for pairs not found in the batch data. From the above discussion, we have that
\begin{equation}
\hat{q}^j_N\left( \xx, a  \right) \approx \max_{\substack{\uu \in \UU\\ \uu(j) = a}} Q_N \left(\xx, \uu \right), \label{eq:approxQ}
\end{equation}
In other words, $\hat{q}^j_N\left( \xx, a  \right)$ can be interpreted as the maximum of the learned, centralized $Q$-function as approximated by agent $j$ when they implement control $a$. Let $\hat{Q}_N^j$ be the approximation of the $Q$-function for agent $j$ after $N$ iterations given the available batch data. 
We can redefine $\hat{q}^j_N\left( \xx, a  \right)$ in terms of the centralized $Q$-function approximation, $\hat{Q}_N^j$, as:
\begin{equation}
\hat{q}^j_N\left( \xx, a  \right) = \max_{\substack{\uu \in \UU\\ \uu(j) = a}} \hat{Q}_N^j \left(\xx, \uu \right).
\label{eq:approx_Qhat}
\end{equation}
The right-hand side of~\eqref{eq:approx_Qhat} is similar to~\eqref{eq:approxQ}'s and, hence, approximates the maximum of the centralized $Q$-function by the \texttt{FQI} approach~\citep{ernst2005tree}.
We assume that $\hat{Q}_N^j$ {\color{blue}is} monotonically increasing. This assumption is justified by the fact that the learned $Q$-function, the $\hat{q}^j$-function, and the \texttt{FQI} approximation of the $Q$-function are all monotonic. The monotonicity follows in all three cases from the structure of the updates when $r(\xx,\uu) \geq 0$ for all $\left(\xx, \uu \right) \in \XX \times \UU$ (see Lemma~\ref{lem:monotonicity}). Thus, we assume that an approximation $\hat{Q}_N^j$ of the centralized $Q$-function from each agent should share this property. 

Next, we extend the aforementioned approach to the stochastic setting. Let $j\in \left\{1,2,\ldots, m\right\}$ and $N \in \NN$. The stochastic analog of~\eqref{eq:d-q}~\citep{lauer2000algorithm} is:
\begin{equation}
q_N^j\left(\xx, a\right) = \begin{cases}
q_{N-1}^j\left(\xx, a\right), \qquad\text{ if } \xx \neq \xx_N \text{ or } a \neq \uu_N(j)\\
\max\left\{ q_{N-1}^j\left(\xx, a\right), \E \left[ r\left(\xx , \uu, \mathbf{w}  \right) \right.\right.\\
\left.\left. + \beta \max_{a' \in \AA^j} q_{N-1}^j\left(f(\xx, \uu, \mathbf{w}), a' \right) \right]\right\},\\
\phantom{q_{N-1}^j\left(\xx, a\right),} \qquad\text{ if } \xx = \xx_N, \uu = \uu_N, \text{ and } a = \uu_N(j).
\end{cases} \label{eq:d-q_stochastic}
\end{equation}
The approximation of the local $q^j_N$-functions for stochastic problems are evaluated as follows. For all $N \in \NN$ and $l=1,2,\ldots, L$, let
\begin{align*}
i^{l,j} &= \left(\xx^l, \uu^l\right), \quad o^{l,j}_N = r^l + \beta \max_{a' \in \AA^j} \hat{q}_{N-1}^j\left(\xx_{+}^l, a' \right),
\end{align*}
where $\hat{q}_{0}^j\left(\xx, a \right) = 0$ for all $\left(\xx, a \right) \in \XX \times \AA$. 
We remark that~\eqref{eq:d-q_stochastic}, in comparison to the deterministic update given in~\eqref{eq:d-q}, requires the evaluation of an expectation when the local $q^j$-function is updated. Hence, the pairs $i^{l,j}$ and $o^{l,j}_N$ cannot be fitted directly as it was done for the deterministic setting.
Similarly to~\citep{ernst2005tree}, a regression tree can be used to estimate an expectation. In our case, we apply a regression tree method over the set of joint states and actions to approximate the expectation from \eqref{eq:d-q_stochastic}'s second line. We refer to this expectation approximation as the local auxiliary $q^j_N$-functions, $\tilde{q}^j_N$, which we express as:
\begin{align}
\tilde{q}^j_N\left( \xx, \uu  \right) &= \sum_{l=1}^L\overline{\text{kernel}}\left(\left(\xx^l,\uu^l\right);(\xx,\uu)\right) o^{l,j}_N,
\label{eq:aux_q}
\end{align}
where $\overline{\text{kernel}}\left(\left(\xx^l,\uu^l\right);(\xx,\uu)\right)$, $l=1,2,\ldots, L$ are computed using a regression tree over the \emph{joint} control set $\UU$. This is motivated by the fact that a regression tree averages the value of the outputs, viz., $r\left(\xx , \uu, \mathbf{w}  \right) + \beta \max_{a' \in \AA^j} q_{N-1}^j\left(f(\xx, \uu, \mathbf{w}), a' \right)$, corresponding to the inputs in a given leaf node or partition, viz., the state-control pairs from the batch data. Alternatively, a linear regression or the Robbins-Monro approximation can be used to estimate the conditional expectation~\citep{ernst2005tree,lauer2000algorithm}. Finally, the approximation of the local $q^j_N$-function at $(\xx, a) \in \XX \times \AA$ is:
\begin{equation}
\begin{aligned}
\hat{q}^j_N\left( \xx, a  \right) = &\sum_{l=1}^L \text{kernel}\left(\left(\xx^l,\uu^l(j)\right);(\xx,a)\right)\\
&\quad\; \cdot \max\left\{\hat{q}_{N-1}^j\left(\xx^l, \uu^l(j) \right), \tilde{q}^j_N\left( \xx^l, \uu^l\right) \right\}, \label{eq:stoch_hatq}
\end{aligned}
\end{equation}
where this time, $\text{kernel}\left(\left(\xx^l,\uu^l (j)\right);(\xx,a)\right)$, $l=1,2,\ldots, L$ are computed using a regression tree over the \emph{local} control space $\AA^j$. We remark that while $\tilde{q}^j_N\left( \xx, \uu  \right)$ is a function of the joint control space, we do need to evaluate its maximum over the joint control space and, therefore, $\tilde{q}^j_N\left( \xx, \uu  \right)$ leads to no scalability issue. Finally, we compute $\hat{q}^j_N$, $N=1,2,\ldots,$ iteratively until a $\left\|\hat{q}^j_N - \hat{q}^j_{N-1} \right\|_\infty < \epsilon$, for some set tolerance $\epsilon > 0$. A detailed representation of \texttt{AMAFQI} is provided in Algorithm~\ref{alg:mafqi}. 
Lastly, establishing an exact relation between the $\hat{q}^j$-functions and the centralized $Q$-function as computed by \texttt{FQI} from Section~\ref{sec:fqi} is a topic for future work.

\begin{algorithm}[tb]
\begin{algorithmic}[1]
\STATEx \textbf{Parameters:} $L$, $\SS_L$, $\beta \in [0, 1)$, $\epsilon > 0$
\STATEx \textbf{Initialization:} $N=0$, $\hat{q}_{0}^j(\xx,a) = 0$ for all $j,\xx, a$. 
\medskip

\STATE Compute $\text{kernel}\left(\left(\xx^l,\uu^l(j)\right);(\xx, \uu(j))\right)$ and $\overline{\text{kernel}}\left(\left(\xx^l,\uu^l\right);(\xx,\uu)\right)$ for all $l$ and $j$ using a regression tree algorithm.
\smallskip

\WHILE{$\left\|\hat{q}^j_N - \hat{q}^j_{N-1} \right\|_\infty \geq \epsilon$}
\STATE $N = N+1$
\FOR{$j = 1,2, \ldots, m$}

\FOR{$l=1,2,\ldots, L$}

\STATE Generate the fitting pairs:
\begin{align*}
i^{l,j} = \left(\xx^l, \uu^l(j) \right), \quad
o^{l,j}_N = r^l + \beta \max_{a' \in \AA} \hat{q}_{N-1}^j\left(\xx_{+}^l, a' \right).
\end{align*}
\ENDFOR
\ENDFOR

\FOR{$j=1,2,\ldots,m$}
\STATE Compute the auxiliary $\tilde{q}^j_N$-function:
\[
\tilde{q}^j_N\left( \xx, \uu  \right) = \sum_{l=1}^L\overline{\text{kernel}}\left(\left(\xx^l,\uu^l\right);(\xx,\uu)\right) o^{l,j}_N.
\]
\STATE Update the $\hat{q}^j_N$-function:
\begin{align*}
\hat{q}^j_N\left( \xx, a  \right) = \sum_{l=1}^L &\text{kernel}\left(\left(\xx^l,\uu^l(j)\right);(\xx,a)\right) \\
&\cdot\max\left\{\hat{q}_{N-1}^j\left(\xx^l, \uu^l(j) \right), \tilde{q}^j_N\left( \xx^l, \uu^l\right) \right\}.
\end{align*}
\ENDFOR

\ENDWHILE

\end{algorithmic}
\caption{Approximated Multi-agent Fitted $Q$ Iteration (\texttt{AMAFQI})}
\label{alg:mafqi}
\end{algorithm}

\begin{algorithm}[tb]
\begin{algorithmic}[1]
\STATEx \textbf{Parameters:} $L$, $\SS_L$, $\beta \in [0, 1)$, $0 < \epsilon \leq \gamma$, $\mathcal{L}(\xx)$ for all $\xx \in \XX$, and $p \in \RR$.
\STATEx \textbf{Initialization:} $N=0$, $\bm{\pi}_{0}(\xx^l) = p\mathbf{1}$ for all $l$.
\medskip

\FOR{all iteration $N$}

\FOR{$\xx \in \XX$}

\FOR{$l$ in $\mathcal{L}(\xx)$}

\STATE Update policy $\bm{\pi}_{N}\left( \xx\right)$ according to~\eqref{eq:policy}.

\ENDFOR

\ENDFOR

\ENDFOR

\IF{$\bm{\pi}(\xx) = p\mathbf{1}$ for $\xx \in \XX$}

\STATE Generalize the greedy policy:
\[
\begin{aligned}
\bm{\hat{\pi}}_N\left( \xx \right) = &\text{ClassificationTree}\left( \left\{\left(\xx^l, \bm{\pi}(\xx^l) \right),\right. \right.\\
&\qquad\qquad\left. \vphantom{\xx^l} \left. \left. l=1,2,\ldots, L \right| \bm{\pi}_N(\xx^l) \neq p\mathbf{1} \right\}, \xx \right)
\end{aligned}
\]

\ENDIF

\end{algorithmic}
\caption{Policy search for \texttt{AMAFQI}}
\label{alg:greedy_pol}
\end{algorithm}

\subsection{Greedy policy search}
\label{sec:policy_search}
Next, we propose a policy search for \texttt{AMAFQI}. We note that there are no guarantees that locally maximizing $\hat{q}^{j}$-functions leads to a joint optimal policy, e.g., if there are many joint optimal controls for a given state, maximizing $\hat{q}^{j}$ across the agents $j=1,2,\ldots,m$ can lead to local controls belonging to different joint optima, thus, resulting in a suboptimal joint control~\citep{lauer2000algorithm}. For this reason, our policy search sequentially identifies controls that yield an increase in $\hat{q}^{j}$'s maximum. The policy search is presented in Algorithm~\ref{alg:greedy_pol} and is shown to be a greedy policy in Theorem~\ref{thm:greedy_policy}.
The search can be extended to decentralized settings using a coordination mechanism~\citep{boutilier1996planning,bu2008comprehensive,vlassis2007concise}. 
Specifically, the decision set is ordered and tie breaking within the regression (Algorithm~\ref{alg:mafqi}, Line 1) or classification (Algorithm~\ref{alg:greedy_pol}, Line 9) trees is done according to this ordering. The batch data is also ordered and made available to all agents. Using this convention, each agent computes the $\hat{q}^{j}$-function for all $j$ and uses Algorithm~\ref{alg:greedy_pol} to compute a unique greedy policy. Because the policy is unique, the local control implemented by agent $j$ leads to a joint greedy control.

Let $j \in \left\{1,2 \ldots, m \right\}$, $l \in \left\{1,2,\ldots, L \right\}$. Let $0 < \epsilon \leq \gamma < +\infty$. 
The parameter $\gamma$ governs how the policy is iteratively updated, and is related to $\epsilon$, the maximum difference between two consecutive $\hat{q}_N^j$ values at convergence (see Algorithm~\ref{alg:mafqi}, line 2).  Parameter $\gamma$ can be equal to but no smaller than $\epsilon$.  Choosing $\gamma$ larger than $\epsilon$ may enable the identification of suboptimal actions in cases where smaller values of $\gamma$ lead to an inconclusive policy search.  In this sense, by enabling a relaxation of the stringency of the policy, this parameter provides practical value, however we leave the exploration of its theoretical properties to future work. Let $\mathcal{L}(\xx) = \left\{\left. l = 1,2,\ldots, L \right| \xx = \xx^l, \left(\xx^l , \uu^l, \xx_{+}^l, r^l \right) \in \SS_L \right\}$ for all $\xx \in \XX$. The set $\mathcal{L}(\xx)$ identifies sample points $l$ such data $\xx^l = \xx$. Let $N \in \NN$ where $N \geq 1$. Consider the policy $\bm{\pi}_{N}:\XX \mapsto \UU$ evaluated at a point from the batch data provided in~\eqref{eq:policy} of page~\pageref{eq:policy} with $\bm{\pi}_{0}\left( \xx\right) = p \mathbf{1}$ for all $\xx \in \XX$. In~\eqref{eq:policy}, $\mathbf{1}$ is an $m$-dimensional vector consisting only of ones and $p$ is an auxiliary parameter used to indicate that no control within the data set corresponds to the greedy maximum for state $\xx$ after the $N^\text{th}$ \texttt{AMAFQI} iteration. It is used to restart the search. If $\bm{\pi}_{N}\left( \xx\right) = p \mathbf{1}$ when the search ends, then the policy for state $\xx$ must be approximated from similar states $\xx'$ for which a greedy decision has been identified, i.e., $\bm{\pi}_{N}\left( \xx'\right) \neq p \mathbf{1}$. This will be discussed at the end of this section. We now have the following results about the policy~\eqref{eq:policy}.

\begin{figure*}[!t]
\begin{equation}
\bm{\pi}_{N}\left( \xx\right) = \begin{cases}
\uu^l, &\text{ if } \max_{a \in \AA^j} \hat{q}^j_{N}\left( \xx, a  \right) - \max_{a \in \AA^j} \hat{q}^j_{N-1}\left( \xx, a  \right) \geq \gamma \ \forall j \in \left\{1,2,\ldots, m \right\}\\
 &\quad \text{ and } \hat{q}^j_{N}\left( \xx, \uu^l(j)  \right) = \max_{a \in \AA^j} \hat{q}^j_{N}\left( \xx, a  \right) \ \forall j \in \left\{1,2,\ldots, m \right\}, \text{ s.t. } l \in \mathcal{L}(\xx)\\
p \mathbf{1}, &\text{ if } \max_{a \in \AA^j} \hat{q}^j_{N}\left( \xx, a  \right) - \max_{a \in \AA^j} \hat{q}^j_{N-1}\left( \xx, a  \right) \geq \gamma \ \forall j \in \left\{1,2,\ldots, m \right\}\\
&\quad \text{ and } \hat{q}^j_{N}\left( \xx, \uu^l(j)  \right) \neq \max_{a \in \AA^j} \hat{q}^j_{N}\left( \xx, a  \right) \text{ for } j \in \left\{1,2,\ldots, m \right\}, \text{ s.t. } l \in \mathcal{L}(\xx)\\
\bm{\pi}_{N-1}\left( \xx\right), &\text{ otherwise. }
\end{cases} \label{eq:policy}
\end{equation}
\hrulefill
\end{figure*}

\begin{theorem}
Let $l \in \left\{1,2, \ldots, L \right\}$ such that $\bm{\pi}_N\left( \xx^l\right)\neq p\mathbf{1}$ and $\overline{\uu} \in \bm{\pi}_N\left( \xx^l\right)$. Then, for all $j = \left\{1,2,\ldots, m\right\}$, we have:
$
\max_{\uu \in \UU} \hat{Q}_N^j \left(\xx^l,\uu \right) - \hat{Q}_N^j \left(\xx^l, \overline{\uu} \right) < 2 \gamma,
$
and $\bm{\pi}_N\left(\xx^l\right)$ is a $2\gamma$-greedy policy at $\xx^l$ with respect to all $\hat{Q}_N^j$, the monotonic approximations of the centralized $Q$-function from each agent.
\label{thm:greedy_policy}
\end{theorem}
\vspace{-0.25cm}

The proof of Theorem~\ref{thm:greedy_policy} is presented in~\ref{app:greedy_policy}.
The above policy search identifies controls using $\hat{q}^j_N$-values that are within $2\gamma$ of $\hat{Q}^j_N$'s maximum for states $\xx$ that belongs to the batch data. This thus leads to a global performance bound provided by the approximation error of the centralized $Q$-function and the $2\gamma$ margin. The search is inconclusive if the optimal control with respect to $\hat{Q}^j_N$ at state $\xx \in \XX$ for some agent $j$ is not in the batch data or if the optimal control performed poorly when sampled to generate the batch data due to stochasticity. 

If $\bm{\pi}(\xx) \neq p\mathbf{1}$ for all $\xx \in \XX$, then the policy can be used directly. If $\bm{\pi}(\xx) = p\mathbf{1}$ for some $\xx \in \XX$, then we use an approximation to generalize the policy to all states similarly to the approach used to generalize the $\hat{q}$-value to all state-control pairs. Let $\bm{\hat{\pi}}_N: \XX \mapsto \UU$ be the approximation of the greedy policy with respect to all $\hat{Q}_N^j$, $j=1,2, \ldots, m$:
$\bm{\hat{\pi}}_N\left( \xx \right) = \text{ClassificationTree}\left( \left\{\left(\xx^l, \bm{\pi}(\xx^l) \right), \left. l=1,2,\ldots, L \right| \bm{\pi}_N(\xx^l) \neq p\mathbf{1} \right\}, \xx \right)$.
Finally, if $\bm{\pi}(\xx) = p\mathbf{1}$ for all $\xx \in \XX$, the batch data does not permit to identify a $2\gamma$-greedy policy with respect to all $\hat{Q}^j_N$-functions.
We remark that $\bm{\hat{\pi}}_N$ only needs to be computed once when the \texttt{AMAFQI} has converged to the $\hat{q}^j$ functions. Thus, a significant advantage of \texttt{AMAFQI}'s policy is that once the \texttt{AMAFQI} algorithm has converged, little to no computations are required to determine the controls when the policy is used. In comparison, the maximum over the joint control space $\UU$ of the approximated $Q$-function needs to be computed when \texttt{FQI} is implemented. This must be done by enumeration because the maximization problem is neither analytically nor numerically solvable. In a multi-agent setting, the joint control space cardinality increases exponentially with the number of agents. Removing the need to compute this maximum reduces the computational burden of \texttt{FQI} when \texttt{AMAFQI} is used.

\subsection{\texttt{AMAFQI-L} update}
\label{ssec:light}
In the previous subsection, we presented a $2\gamma$-greedy policy search with respect to the approximations of the centralized $Q$-function of all agents $j$. This policy search can be modified to only use the $\hat{q}^j_N$-function of a single agent $j$. We refer to this alternate policy as \texttt{AMAFQI-L}. Because of~\eqref{eq:approxQ}, the maximum of a single $\hat{q}^j_N$ still approximates the centralized $Q$-function's maximum. The difference is that \texttt{AMAFQI-L} is now a $2\gamma$-greedy policy with respect to agent $j$'s approximation of the centralized $Q$-function rather than with respect to the approximation of all agents. Thus, this approximation is looser than the previous one. The main gain is, however, computational efficiency because only a single $\hat{q}^j$-function must be iteratively computed. The computational requirement is constant with respect to the number of agents whereas it scales linearly and exponentially with the number of agents for \texttt{AMAFQI} and \texttt{FQI}, respectively.

\texttt{AMAFQI-L}'s algorithm is similar to \texttt{AMAFQI}, except that $j$ is set to a constant value within $\left\{1,2, \ldots, m \right\}$ throughout the iterations $N$ and the policy search. The algorithm is presented in Algorithm~\ref{alg:mafqi_l} of~\ref{app:amafqi_l}. The \texttt{AMAFQI-L} policy search is identical to \texttt{AMAFQI}'s, but uses~\eqref{eq:policy_l} instead of~\eqref{eq:policy}:
\begin{equation}
\bm{\pi}_{N}^\texttt{L}\left( \xx\right) = \begin{cases}
\uu^l, &\hspace{-0.8cm}\text{ if } \max_{a \in \AA^j} \hat{q}^j_{N}\left( \xx, a  \right) - \max_{a \in \AA^j} \hat{q}^j_{N-1}\left( \xx, a  \right) \geq \gamma \\
 &\hspace{-1.4cm}\text{ and } \hat{q}^j_{N}\left( \xx, \uu^l(j)  \right) = \max_{a \in \AA^j} \hat{q}^j_{N}\left( \xx, a  \right), \text{ s.t. } l \in \mathcal{L}(\xx)\\
p \mathbf{1}, &\hspace{-0.8cm}\text{ if } \max_{a \in \AA^j} \hat{q}^j_{N}\left( \xx, a  \right) - \max_{a \in \AA^j} \hat{q}^j_{N-1}\left( \xx, a  \right) \geq \gamma\\
&\hspace{-1.4cm}\text{ and } \hat{q}^j_{N}\left( \xx, \uu^l(j)  \right) \neq \max_{a \in \AA^j} \hat{q}^j_{N}\left( \xx, a  \right), \text{ s.t. } l \in \mathcal{L}(\xx)\\
\bm{\pi}_{N-1}\left( \xx\right), &\hspace{-0.4cm}\text{ otherwise. }
\end{cases}\label{eq:policy_l}
\end{equation}
The above discussion is formalized by the following result.
\begin{corollary}
Consider Theorem~\ref{thm:greedy_policy}'s assumptions, and suppose $j \in \left\{1,2,\ldots, m\right\}$. If $\overline{\uu} \in \bm{\pi}_N^\texttt{L}\texttt{}\left( \xx^l\right)$, then we obtain:
% \[
$\max_{\uu \in \UU} \hat{Q}_N^j \left(\xx^l,\uu \right) - \hat{Q}_N^j \left(\xx^l, \overline{\uu} \right) < 2 \gamma,$
% \]
and $\bm{\pi}_N^\texttt{L}\left(\xx^l\right)$ is a $2\gamma$-greedy policy at $\xx^l$ with respect to $\hat{Q}_N^j$.
\label{thm:greedy_policy_L}
\end{corollary}
The proof follows from Theorem~\ref{thm:greedy_policy} for a single $j$.

\section{Convergence}
\label{sec:convergence}
We show that each local $\hat{q}^j_N$-function defined in~\eqref{eq:stoch_hatq} converges to a unique and finite function with respect to the infinity norm. We first establish the monotonicity of $\hat{q}^j_N$ for all $j$.

\begin{lemma}
Suppose $r\left(\xx, \uu, \mathbf{w} \right) \geq 0$ and $\hat{q}_0^j\left(\xx, a \right)=0$ for all $(\xx, a, \mathbf{w}) \in \XX \times \AA \times \mathcal{W}$, then $\hat{q}_{N}^j\left(\xx,a\right) \leq \hat{q}_{N+1}^j\left(\xx,a\right)$ for all $(\xx, a) \in \XX \times \AA$ and $N \in \NN$.
\label{lem:monotonicity}
\end{lemma}
\vspace{-0.25cm} 

We now state the convergence result.

\begin{theorem}
Suppose Assumptions~1$-$3 hold and $\hat{q}^j_0 (\xx, a) = 0$ for all $(\xx, a, \mathbf{w}) \in \XX \times \AA \times \mathcal{W}$ and $j=1,2,\ldots, m$. Then $\hat{q}^j_N\left(\xx, a \right)$ converges to the unique limit $\hat{q}^j_{\mathcal{S}_L}\left(\xx, a \right)$, i.e., the unique maximum of the {$\hat{Q}^j$-function} for $\xx$ and $\uu(j) = a$ when estimated using the data set $\mathcal{S}_L$ and a regression tree method. 
Moreover, for all $\epsilon >0$, there exists $n(j) \in \mathbb{N}$ such that for all $N \geq n(j)$, 
\[
\left\| \hat{q}^j_N  -  \hat{q}^j_{\mathcal{S}_L} \right\|_{\infty} < \epsilon.
\]
\label{thm:convergence}
\end{theorem}
\vspace{-0.5cm}
The proofs of Lemma~\ref{lem:monotonicity} and Theorem~\ref{thm:convergence} are in~\ref{app:lem_mono} and~\ref{app:thm_conv}, respectively.
Theorem~\ref{thm:convergence} ensures that there exist unique, finite-valued $\hat{q}^j$-functions for the data set $\mathcal{S}_L$ which can be used for the policy search. Thus, $\hat{q}^j_{\mathcal{S}_L}$-functions for the data set $\mathcal{S}_L$ can always be computed under the aforementioned assumptions. We remark that Theorem~\ref{thm:convergence} applies to \texttt{AMAFQI} and \texttt{AMAFQI-L} because it holds for any $j$.

Similarly to~\citep{ernst2005tree}, the error due to the regression-tree method (or any other supervised learning approach) is not modeled explicitly in this work. For \texttt{AMAFQI}, this error would translate in $\hat{Q}^j$-functions suffering itself from a larger approximation error. We remark that using a regression tree method allows us to establish the \texttt{AMAFQI}'s convergence. The regression error is a topic for future investigation.

\section{Numerical examples}
\label{sec:num}
In this section, we compare the performance of \texttt{AMAFQI} and \texttt{FQI} in numerical simulations. Our comparison uses \texttt{FQI} because it provides a learned $Q$-function that is the unique solution to Bellman's equation given the batch data~\citep{ernst2005tree}. It can, therefore, be considered as an adequate benchmark for the batch reinforcement learning setting. We test our approach on a multi-agent, multi-state random problem similar to the example presented in~\citep{zhang2018fully,dann2014policy}.

Let $\hat{Q}^\texttt{FQI}_N: \XX \times \UU \mapsto \RR$ be the approximated $Q$-function after $N$ iterations evaluated via \texttt{FQI}~\citep{ernst2005tree}. Single problem instance simulations are run on a $2.4$ GHz Intel Core i$5$ laptop computer and multiple instance simulations are run on the Savio computational cluster resource from the Berkeley Research Computing program. The computations of $\hat{q}^j_N$ and $\hat{Q}^\texttt{FQI}_N$ for all samples $l$ are parallelized to reduce the full computation time. 

\subsection{Setting}
The multi-agent, multi-state random problem is as follows. We consider $m$ agent having to coordinate their individual binary decision to reach one of the $X$ joint states and maximize their reward over $\tau$ rounds. The joint binary decision determines the probability of moving from one state to another. Let $P(\xx):\UU\times \XX\mapsto \XX$ be the transition matrix for state $\xx \in \XX$. All transition matrices are randomly generated according to uniform distributions and then normalized to obtain row-stochastic matrices. The reward is determined by the joint state at the end of a round. Let the mean reward for a state $\xx \in \XX$ be $R(\xx) \sim \text{Uniform[0,5]}$. The reward for reaching state $\xx \in \XX$ is then $r(\xx) \sim \text{Uniform}[R(\xx) - \frac{1}{2}, R(\xx) + \frac{1}{2}]$.

\subsection{Experiments}

We use \texttt{Totally Randomized Trees}~\citep{geurts2006extremely} for the regression tree. We consider ensembles of $5$ trees with each at a minimum of $10$ data points in a leaf node. We let $\beta=0.5$.

\subsubsection{5 agents}

\begin{figure}[t!]
    \centering
    \vspace{-0.5cm}
    \includegraphics[width=1\columnwidth]{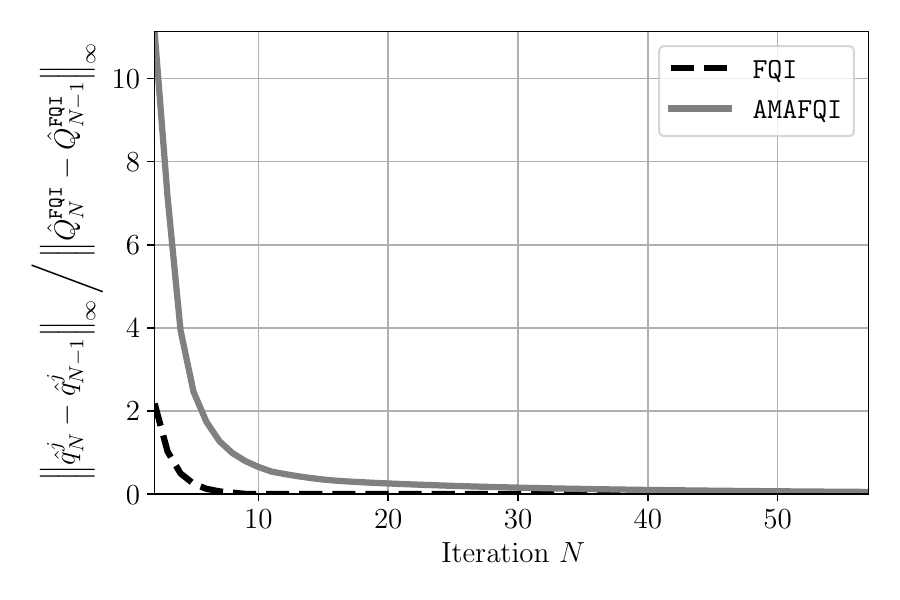}
    \vspace{-1cm}
    \caption{Convergence of \texttt{AMAFQI} and \texttt{FQI} in the $5$-player, $5$-state problem}
    \label{fig:conv_multi}
\end{figure}

We let $m=5$ and $\card \XX = 5$. We uniformly sample $L= 2000$ $\left(\xx^l,\uu^l, \xx_{+}^l, r^l \right)$-tuples. The convergence of both \texttt{AMAFQI} and \texttt{FQI} implementations for this numerical experiment is shown in Figure~\ref{fig:conv_multi}. Figure~\ref{fig:conv_multi} shows that $\left\| \hat{q}^j_N - \hat{q}^j_{N-1} \right\|_{\infty} $ and $\left\| \hat{Q}^\texttt{FQI}_N - \hat{Q}^\texttt{FQI}_{N-1} \right\|_{\infty}$ go to zero as $N$ increases. Thus, both values converge to their respective unique and finite limits.

\begin{figure}[t!]
    \centering
    \vspace{-0.5cm}
    \includegraphics[width=1\columnwidth]{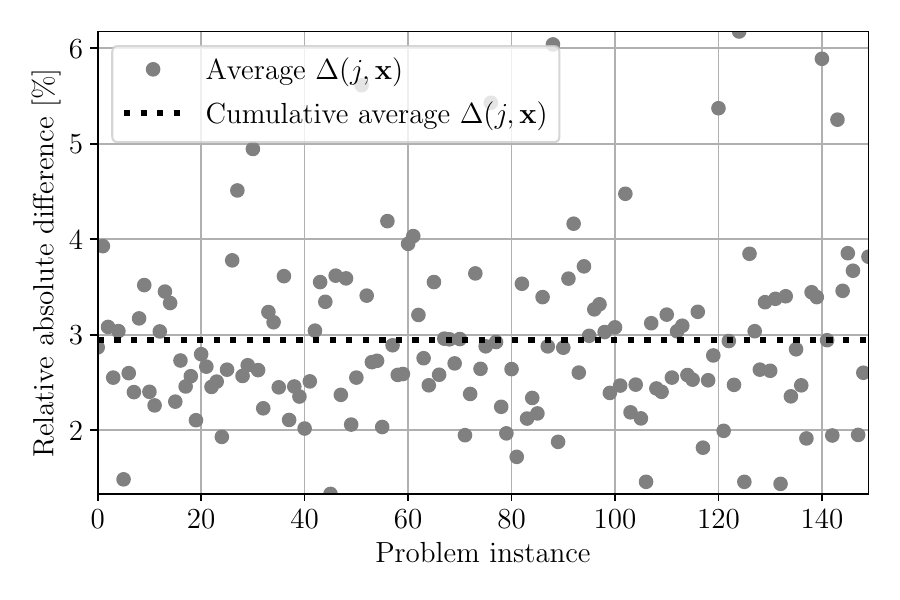}
    \vspace{-1cm}
    \caption{Average $\Delta(j,\xx)$ over all $j,\xx$ for $150$ random instances of the 5-agent, 5-state problem}
    \label{fig:diff_max_dataset_multi}
\end{figure}

We compare the approximated value function at $\xx$ for \texttt{AMAFQI} and \texttt{FQI} using the relative absolute difference between both maxima, defined as $\Delta(j,\xx) = \left|\frac{\max_{a \in \AA} \hat{q}^j_N\left( \xx, a\right) - \max_{\uu \in \UU} \hat{Q}^\texttt{FQI}_N \left( \xx , \uu \right)}{\max_{\uu \in \UU} \hat{Q}_N^\texttt{FQI} \left( \xx , \uu \right)} \right|$, for $j=1,2,\ldots, m$ and $\xx \in \XX$. We sequentially compute the $\hat{q}^j$- and $\hat{Q}^\texttt{FQI}$-functions for $150$ different problem instances, each time sampling a new data set $\SS_L$. The average $\Delta(j,\xx)$ for all the problem instances are reported in Figure~\ref{fig:diff_max_dataset_multi}. The average over all problem instances of the relative difference $\Delta(j,\xx)$ is $2.92\%$.

\begin{figure}[t!]
    \centering
    \vspace{-0.5cm}
    \includegraphics[width=1\columnwidth]{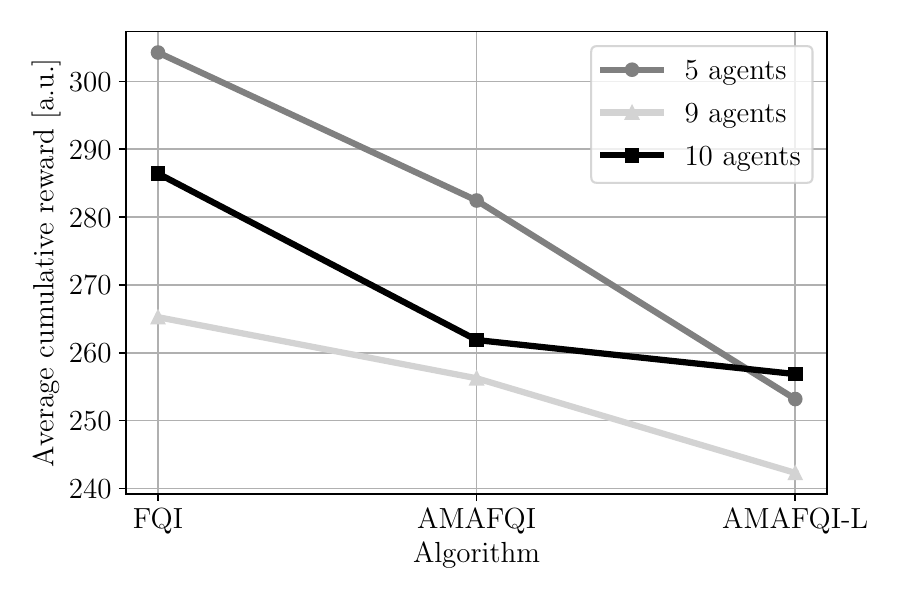}
    \vspace{-1cm}
    \caption{Average cumulative reward for the $5$, $9$, and $10$-agent, $5$-state problem over $150$, $10$, and $5$ problem instances, respectively}
    \label{fig:reward_comp}
\end{figure}

For each problem instance, we compute the reward obtained by the greedy policies over $100$ trials each with a time horizon $\tau = 100$ rounds. For each trial, the initial state is randomly sampled. The average reward of \texttt{FQI}'s, \texttt{AMAFQI}'s, and \texttt{AMAFQI-L}'s greedy policies are shown in Figure~\ref{fig:reward_comp}. The relative difference in average cumulative reward between \texttt{AMAFQI} and \texttt{FQI} is small and only $7.17\%$. The performance of \texttt{AMAFQI-L} is lower than \texttt{AMAFQI}'s and leads to a $16.79\%$ cumulative reward decrease in comparison to \texttt{FQI}.

We conclude by discussing the computation time of \texttt{AMAFQI}. The average computation time for a single iteration $N$ and until convergence for \texttt{FQI}, \texttt{AMAFQI} and \texttt{AMAFQI-L} are reported in Table~\ref{tab:avg_time_multi} for the $150$ problem instances. The numbers from Table~\ref{tab:avg_time_multi} given in parentheses and the subsequent similar tables represent the total computation times which includes the policy search. An iteration of \texttt{AMAFQI} and \texttt{AMAFQI-L} with and without the policy search has a shorter duration than an \texttt{FQI} iteration. Because the approximation requires more $N$ iterations, \texttt{AMAFQI} still takes more time to converge. The amount of time to convergence for \texttt{AMAFQI-L} and \texttt{FQI} are similar. The problem size is still small given its binary controls and only $5$ agents. Hence, an approach tailored to multi-agent settings is not necessarily needed yet. We provide this example of a small problem instance so that both \texttt{AMAFQI} and \texttt{FQI} can be simulated repetitively in an acceptable time frame. The comparison's bottleneck is \texttt{FQI} which is computationally very time consuming. Thus, given our computing infrastructure, we restrict our analysis to $10$ agents or less as our objective is to compare \texttt{AMAFQI}'s performances to \texttt{FQI}'s on several instances.

\begin{table}[t!]
    \caption{Average computation times for the $5$-agent, $5$-state problem ($150$ problem instances, $100$ trials)}
    \label{tab:avg_time_multi}
    \centering

    \begin{tabular}{c cc}
    \hline

    \hline
    \textbf{Average time} & \textbf{Iteration [s]} & \textbf{Convergence (policy) [s]}\\
    \hline
    \texttt{FQI}     & $23.39$ & $155.00$ \\
    \texttt{AMAFQI}  & $12.09$ & $577.20$ ($658.45$) \\
    \texttt{AMAFQI-L}    & $2.41$ & $115.44$ ($135.11$)\\
    \hline

    \hline
    \end{tabular}
\end{table}

\subsubsection{9 and 10 agents}

When the number of agents increases, the computational advantage of \texttt{AMAFQI} is clear. Tables~\ref{tab:avg_time_multi9} and~\ref{tab:avg_time_multi10} present the computation times for $m=9$ with $L=5000$ and $m=10$ with $L=7000$, respectively. The average $\Delta(j,\xx)$ is $8.17\%$ when $m=9$ and $7.90\%$ when $m=10$. We note that $\Delta(j,\xx)$ can be further reduced by increasing $L$ at the expense of a longer computation time. The averaged cumulative reward for the $100$ trials of each problem instance is provided in Figure~\ref{fig:reward_comp} for both the $9$- and $10$-agent problem.

As shown in Tables~\ref{tab:avg_time_multi9} and~\ref{tab:avg_time_multi10}, \texttt{AMAFQI} requires much less computation time than \texttt{FQI} to converge when $m$ increases and only leads to a limited decrease in cumulative reward. In the present case, we register a $3.40\%$ ($m=9$) and $8.57\%$ ($m=10$) reduction of the average reward when using \texttt{AMAFQI}. Moreover, for \texttt{AMAFQI}, the total computation time until convergence includes most of the calculations required for the evaluation step. \texttt{AMAFQI-L} further reduces the total computation time.
For $m=9$, \texttt{AMAFQI-L} requires less than $8$ minutes to convergence and to compute the policy instead of $84$ minutes for \texttt{AMAFQI} and $3$ hours ($177$ minutes) for \texttt{FQI}. When considering $m=10$, \texttt{AMAFQI-L} needs $14$ minutes whereas \texttt{AMAFQI} and \texttt{FQI} takes, respectively, $3$ hours ($181$ minutes) and $12$ hours ($723$ minutes). The performance of the \texttt{AMAFQI-L} policy is slightly lower and leads to a decrease in the cumulative reward of $8.65\%$ ($m=9$) and $10.32\%$ ($m=10$) with respect to \texttt{FQI}.

\begin{table}[t!]
    \caption{Average computation times for the $9$-agent, $5$-state problem ($10$ problem instances)}
    \label{tab:avg_time_multi9}
    \centering

    \begin{tabular}{c cc}
    \hline

    \hline
    \textbf{Average time} & \textbf{Iteration [s]} & \textbf{Convergence (policy) [s]}\\
    \hline
    \texttt{FQI}     & $1660.07$ & $10615.95$ \\
    \texttt{AMAFQI}  & $77.31$ & $3766.03$ $(4998.52)$ \\
    \texttt{AMAFQI-L}    & $8.59$ & $418.44$ $(454.23)$\\
    \hline

    \hline
    \end{tabular}
\end{table}

\begin{table}[t!]
    \caption{Average computation times for the $10$-agent, $5$-state problem ($5$ problem instances)}
    \label{tab:avg_time_multi10}
    \centering

    \begin{tabular}{c cc}
    \hline

    \hline
    \textbf{Average time} & \textbf{Iteration [s]} & \textbf{Convergence (policy) [s]}\\
    \hline
    \texttt{FQI}     & $6579.77$ & $43421.90$ \\
    \texttt{AMAFQI}  & $156.58$ & $7859.76$ ($10840.89$) \\
    \texttt{AMAFQI-L}    & $15.67$ & $785.98$ ($785.98$)\\
    \hline

    \hline
    \end{tabular}
\end{table}

\section{Conclusion}
\label{sec:conclu}

In this work, we propose the \texttt{AMAFQI} algorithm, a tractable multi-agent approximation of \texttt{FQI} for batch reinforcement learning problems. We design an iterative policy search for \texttt{AMAFQI} and demonstrate that it is a greedy policy with respect to an approximation of the learned $Q$-function of all agents. Our approach performs computations only over local control sets contrarily to \texttt{FQI} that works over the joint control space. The number of calculations required in each iteration of the algorithm grows linearly and exponentially with the number of agents, respectively, for \texttt{AMAFQI} and for \texttt{FQI}. Consequently, \texttt{FQI} is impractical and quickly intractable in presence of multiple agents. Our approach offers an efficient alternative for multi-agent batch reinforcement learning problems. We present a derivative of our approach, \texttt{AMAFQI-L}, which further reduces the computational burden of \texttt{AMAFQI}.

We consider a multi-agent batch reinforcement learning problem and compare the performance of \texttt{AMAFQI} with \texttt{FQI}. Numerical simulations show that the value functions computed by our approximation and by \texttt{FQI} are similar, e.g., with a discrepancy of $2.92\%$ when $m=5$, and that the performance level is also alike, e.g., with a difference of $7.12\%$. Lastly, computation times are compared and \texttt{AMAFQI} and \texttt{AMAFQI-L} outperform significantly \texttt{FQI} when the number of agent increases. For example, \texttt{AMAFQI} and \texttt{AMAFQI-L} require, respectively, only $181$ minutes and $13$ minutes against a total computation time of $723$ minutes, on average, for \texttt{FQI} when $m=10$.

In future work, we wish wish to investigate approaches to reduce the number of $N$ iterations performed in \texttt{AMAFQI} before convergence, e.g., by considering the growing batch learning paradigm~\citep{lange2012batch} in which an exploration policy is used, and new observed transitions are periodically incorporated in the batch data. Lastly, we would like to compare \texttt{AMAFQI} to \texttt{FQI} in a more sophisticated setting and use \texttt{AMAFQI} to dispatch flexible loads for network-safe demand response in unknown electric grids. This is a topic for future work.

\section*{Acknowledgements}
This work was funded in part by the Institute for Data Valorization (IVADO), in part by the Natural Sciences and Engineering Research Council of Canada, in part by the National Science Foundation, award 1351900, and in part by the Advanced Research Projects Agency-Energy, award DE-AR0001061.

This research used the Savio computational cluster resource provided by the Berkeley Research Computing program at the University of California, Berkeley (supported by the UC Berkeley Chancellor, Vice Chancellor for Research, and Chief Information Officer).

\appendix

\section{Proof of Theorem~\ref{thm:greedy_policy}}
\label{app:greedy_policy}

We base our proof on~\citep[Proposition 2]{lauer2000algorithm}. Consider the monotonic approximation of the centralized $Q$-functions from all agents, $\hat{Q}^j_N$, $j=1,2,\ldots, m$. Let $l \in \left\{1,2, \ldots, L \right\}$. Let $0 \leq N' < N$ such that for all $j \in \left\{1,2,\ldots, m \right\}$ we have:
\begin{equation}
\max_{\uu \in \UU} \hat{Q}_{N'+1}^j\left( \xx^l , \uu \right) - \max_{\uu \in \UU} \hat{Q}^j_{N'}\left( \xx^l , \uu \right) \geq \gamma, 
\label{eq:Qhat_smaller}
\end{equation}
and,
\begin{equation}
\max_{\uu \in \UU} \hat{Q}^j_{n}\left( \xx^l , \uu \right) - \max_{\uu \in \UU} \hat{Q}^j_{N'+1}\left( \xx^l , \uu \right) < \gamma, 
\label{eq:Qhat_equal}
\end{equation}
for $n = N' +2, N'+3, \ldots, N$. From the approximation definition~\eqref{eq:approx_Qhat}, we equivalently have for all $j\in \left\{1,2,\ldots,m \right\}$:
\begin{equation}
\max_{a \in \AA^j} \hat{q}_{N'+1}^j\left( \xx^l , a \right) - \max_{a \in \AA^j} \hat{q}_{N'}^j\left( \xx^l , a \right) \geq \gamma, 
\label{eq:qhat_smaller}
\end{equation}
and
\begin{equation}
\max_{a \in \AA^j} \hat{q}_{n}^j\left( \xx^l , a \right) - \max_{a \in \AA^j} \hat{q}_{N'+1}^j\left( \xx^l , a \right) < \gamma, 
\label{eq:qhat_equal}
\end{equation}
for $n = N' +2, N'+3, \ldots, N$. 
By~\eqref{eq:qhat_smaller} and~\eqref{eq:qhat_equal}, the last update to the policy at $\xx^l$ can only occur at $N'+1$. Regarding the policy update, if  $\hat{q}^j_{N+1}\left( \xx, \uu^l(j)  \right) = \max_{a \in \AA^j} \hat{q}^j_{N+1}\left( \xx, a  \right)$ such that $l \in \mathcal{L}(\xx)$ for all $j$, then this last update was performed when the control $\uu^l$ was considered by the \texttt{AMAFQI} update. Otherwise, if there exists no $l \in \mathcal{L}(\xx)$ such that $\hat{q}^j_{N+1}\left( \xx, \uu^l(j)  \right) = \max_{a \in \AA^j} \hat{q}^j_{N+1}\left( \xx, a  \right)$ or the equality does not hold for all $j$, the search is inconclusive for the iteration $N$. By assumption, $\bm{\pi}_{N'+1}(\xx^l) \neq p\mathbf{1}$ and at least one policy update was performed.

Finally, iteration $N'+1$ coincides to the last time the maximum $\hat{Q}^j$-function changed by at least $\gamma$ for all $j$ because of~\eqref{eq:Qhat_smaller} and~\eqref{eq:Qhat_equal}. Thus, for all $\overline{\uu}_{N'+1} \in \bm{\pi}_{N'+1}(\xx^l)$ we have
\begin{equation}
\max_{\uu \in \UU}\hat{Q}^j_{N' + 1 } \left( \xx^l ,\uu \right) - \hat{Q}^j_{N' + 1 } \left( \xx^l , \overline{\uu}_{N'+1} \right) < \gamma,
\label{eq:withPi}
\end{equation}
for all $j \in \left\{1,2,\ldots,m \right\}$. The monotonicity of the $\hat{Q}^j_N$-function implies that~\eqref{eq:withPi} can be re-expressed as
\begin{equation}
\max_{\uu \in \UU}\hat{Q}_{N' + 1 }^j \left( \xx^l ,\uu \right) - \hat{Q}_{N}^j \left( \xx^l , \overline{\uu}_{N'+1} \right) < \gamma.
\label{eq:following_mono}
\end{equation}
From~\eqref{eq:Qhat_equal}, we know that
\begin{equation}
\max_{\uu \in \UU} \hat{Q}_N^j\left(\xx^l, \uu \right) - \gamma < \max_{\uu \in \UU} \hat{Q}_{N'+1}^j\left(\xx^l, \uu \right).
\label{eq:using}
\end{equation}
Using~\eqref{eq:using} in~\eqref{eq:following_mono}, we obtain $\max_{\uu \in \UU}\hat{Q}_{N }^j \left( \xx^l ,\uu \right) - \hat{Q}_{N}^j \left( \xx^l , \overline{\uu}_{N'+1} \right) < 2\gamma$.
Lastly, because the policy is not updated between $N' +1$ and $N$, we have that $\bm{\pi}_{N'+1} \left(\xx^l \right) = \bm{\pi}_N \left(\xx^l \right)$ and thus, we have $\max_{\uu \in \UU}\hat{Q}_{N }^j \left( \xx^l ,\uu \right) - \hat{Q}_{N}^j \left( \xx^l , \overline{\uu}_{N} \right) < 2\gamma$,
where $\overline{\uu}_{N} \in \bm{\pi}_N \left(\xx^l \right)$. Hence, the policy $\bm{\pi}_N\left(\xx^l\right) \neq p \mathbf{1}$ is a $2\gamma$-greedy policy for the approximation of the centralized $Q$-function of all agents.$\qedproof$

\section{\texttt{AMAFQI-L} algorithm}
\label{app:amafqi_l}

\begin{algorithm}[H]
\begin{algorithmic}[1]

\STATEx \textbf{Parameters:} $L$, $\SS_L$, $\beta \in [0, 1)$, $\epsilon > 0$, $j \in \left\{1,2, \ldots, m \right\}$.
\STATEx \textbf{Initialization:} $N=0$, $\hat{q}_{0}^j(\xx,a) = 0$ for all $\xx, a$. 
\medskip

\STATE Compute $\text{kernel}\left(\left(\xx^l,\uu^l(j)\right);(\xx, \uu(j))\right)$ and $\overline{\text{kernel}}\left(\left(\xx^l,\uu^l\right);(\xx,\uu)\right)$ for all $l$ using a regression tree algorithm.
\smallskip

\WHILE{$\left\|\hat{q}^j_N - \hat{q}^j_{N-1} \right\|_\infty \geq \epsilon$}
\STATE $N = N+1$

\FOR{$l=1,2,\ldots, L$}

\STATE Generate the fitting pairs:
\begin{align*}
i^{l,j} &= \left(\xx^l, \uu^l(j) \right)\\
o^{l,j}_N &= r^l + \beta \max_{a' \in \AA} \hat{q}_{N-1}^j\left(\xx_{+}^l, a' \right).
\end{align*}
\ENDFOR

\STATE Compute the auxiliary $\tilde{q}^j_N$-function:
\[
\tilde{q}^j_N\left( \xx, \uu  \right) = \sum_{l=1}^L\overline{\text{kernel}}\left(\left(\xx^l,\uu^l\right);(\xx,\uu)\right) o^{l,j}_N.
\]
\STATE Update the $\hat{q}^j_N$-function:
\begin{align*}
\hat{q}^j_N\left( \xx, a  \right) = \sum_{l=1}^L &\text{kernel}\left(\left(\xx^l,\uu^l(j)\right);(\xx,a)\right) \\
&\cdot\max\left\{\hat{q}_{N-1}^j\left(\xx^l, \uu^l(j) \right), \tilde{q}^j_N\left( \xx^l, \uu^l\right) \right\}.
\end{align*}

\ENDWHILE

\end{algorithmic}
\caption{Approximated Multi-agent Fitted $Q$ Iteration -- Light (\texttt{AMAFQI-L})}
\label{alg:mafqi_l}
\end{algorithm}

\section{Proof of Lemma~\ref{lem:monotonicity}}
\label{app:lem_mono}

We prove this lemma by induction. Let $(\xx, a) \in \XX \times \AA$ and $j \in \left\{1,2,\ldots, m \right\}$. For $N=0$, we have $\hat{q}_0^j\left(\xx, a \right) = 0$ for all $\xx, a$ by assumption. For $N=1$, we then have:
\begin{align*}
\hat{q}_1^j\left(\xx, a \right) &= \sum_{l=1}^L \text{kernel}\left(\left(\xx^l,\uu^l(j)\right);(\xx,a)\right) \\
&\qquad\qquad\qquad \cdot \max\left\{\hat{q}_{0}^j\left(\xx^l, \uu^l(j) \right), \tilde{q}^j_1\left( \xx^l, \uu^l\right) \right\}\\
&= \sum_{l=1}^L \text{kernel}\left(\left(\xx^l,\uu^l(j)\right);(\xx,a)\right) \max\left\{0,  r^l\right\}
\end{align*}
because $\sum_{l=1}^L\overline{\text{kernel}}\left(\left(\xx^l,\uu^l\right);(\xx,\uu)\right)=1$ for all $\left(\xx, \uu \right) \in \XX \times \UU$. By assumption, $r^l \geq 0$ and, therefore, $\hat{q}_0^j\left(\xx, a \right) \leq \hat{q}_1^j\left(\xx, a \right)$.
We now show that, the induction hypothesis, $\hat{q}_{N}^j\left(\xx,a\right) \leq \hat{q}_{N+1}^j\left(\xx,a\right)$, holds for $N \rightarrow N+1$. At $N+1$, the $\hat{q}^j$-function is
\begin{align}
\hat{q}_{N+1}^j\left(\xx, a \right) = \sum_{l=1}^L &\text{kernel}\left(\left(\xx^l,\uu^l(j)\right);(\xx,a)\right) \label{eq:hat_tobound}\\
& \cdot\max\left\{\hat{q}_N^j\left(\xx^l, \uu^l(j) \right), \tilde{q}_{N+1}^j\left(\xx^l,\uu^l \right) \right\},\nonumber
\end{align}
where
\begin{equation}
\begin{aligned}
\tilde{q}^j_{N+1}(\xx, \uu)= \sum_{l=1}^L\overline{\text{kernel}}&\left(\left(\xx^l,\uu^l\right);(\xx,\uu)\right)\\
&\cdot \left[ r^l + \beta \max_{a' \in \AA} \hat{q}_{N}^j\left(\xx_{+}^l, a' \right) \right] 
\end{aligned}
\label{eq:tilde_tobound}
\end{equation}
We first use the induction hypothesis in~\eqref{eq:tilde_tobound} and obtain
\begin{align}
\tilde{q}^j_{N+1}(\xx, \uu) &\leq \sum_{l=1}^L\overline{\text{kernel}}\left(\left(\xx^l,\uu^l\right);(\xx,\uu)\right)\nonumber\\
&\qquad\qquad\qquad \cdot \left[ r^l + \beta \max_{a' \in \AA} \hat{q}_{N+1}^j\left(\xx_{+}^l, a' \right) \right]\nonumber\\
&\leq \tilde{q}^j_{N+2}(\xx, \uu) \label{eq:mono_tilde}
\end{align}
Second, we use the induction hypothesis and~\eqref{eq:mono_tilde} in~\eqref{eq:hat_tobound}. This leads to
\begin{align*}
\hat{q}_{N+1}^j\left(\xx, a \right) &\leq \sum_{l=1}^L \text{kernel}\left(\left(\xx^l,\uu^l(j)\right);(\xx,a)\right)\\
&\qquad\quad \cdot \max\left\{\hat{q}_{N+1}^j\left(\xx^l, \uu^l(j) \right), \tilde{q}_{N+2}^j\left(\xx^l,\uu^l \right) \right\} \\
&=\hat{q}_{N+2}^j\left(\xx, a \right)
\end{align*}
where we last used the definition of $\hat{q}_{N+2}^j$. Thus, we have established that $\hat{q}_N^j\left(\xx, a\right)$  is monotonically increasing for all $(\xx, a) \in \XX \times \AA$ and all $N \in \NN$. $\qedproof$

\section{Proof of Theorem~\ref{thm:convergence}}
\label{app:thm_conv}

We first show that $\hat{q}_N^j$ is bounded. By Assumption~\ref{ass:reward}, we have $r(\xx, \uu, \mathbf{w}) \leq R$. Let $j \in \left\{1,2,\ldots, m\right\}$. By definition, $\hat{q}^j_0(\xx, a) = 0$ for all $(\xx, \uu) \in \XX \times \UU$. For $N=1$, we have
\begin{align*}
\left\|\hat{q}_{1}^j\left(\xx, a \right) \right\|_{\infty} &\leq \left\|\sum_{l=1}^L \text{kernel}\left(\left(\xx^l,\uu^l(j)\right);(\xx,a)\right) \right.\\
&\left. \cdot\max\left\{0, \sum_{l=1}^L\overline{\text{kernel}}\left(\left(\xx^l,\uu^l\right);(\xx^l,\uu^l)\right) R \right\}\right\|_\infty\\
&=\max\left\{0,  R \right\}
\end{align*}
because kernels are non-negative and their sum is normalized. By the same process, we sequentially bound $\hat{q}_{N}^j\left(\xx, a \right)$ for all $N \in \NN$:
\begin{align}
\left\|\hat{q}_{N}^j\left(\xx, a \right)\right\|_{\infty} &\leq \left\|\sum_{l=1}^L \text{kernel}\left(\left(\xx^l,\uu^l(j)\right);(\xx,a)\right)\right. \label{eq:last_bound} \\
& \quad \cdot \left.\max\left\{\sum^{N-1}_{n=1} \beta^{n-1}R, R + \beta \sum^{N-1}_{n=1} \beta^{n-1} R \right\}\right\|_\infty 
\nonumber
\end{align}
We further bound~\eqref{eq:last_bound} and obtain: $\left\|\hat{q}_{N}^j\left(\xx, a \right)\right\|_\infty \leq \frac{R}{1-\beta}$ for all $N \in \NN$.
Therefore, $\left\|\hat{q}^j_N\left(\xx, a \right) \right\|_{\infty}$ is bounded from above for all $j \in \{1,2,\ldots,m\}$, and $N \in \NN$. We remark that this is an upper bound and not necessarily the supremum of $\hat{q}_N^j$.

By the monotone convergence theorem, $\hat{q}_N^j\left(\xx, a \right) \rightarrow \hat{q}^j_{\mathcal{S}_L}\left(\xx, a \right)$, where $\hat{q}^j_{\mathcal{S}_L}\left(\xx, a \right) \leq \frac{R}{1-\beta}$ is the supremum of the sequence given in~\eqref{eq:stoch_hatq} at $\left(\xx, a \right)$ because the sequence is monotonically increasing by Lemma~\ref{lem:monotonicity} and is bounded from above. A limit is unique if it exists and therefore $\hat{q}^j_{\mathcal{S}_L}\left(\xx, a \right)$ is the unique solution of~\eqref{eq:q_hat} at $\left(\xx, a \right) \in \XX \times \AA$ given the data set ${\mathcal{S}_L}$. It follows from~\eqref{eq:approx_Qhat} that the limit is the maximum of the centralized $Q$-function approximation at $\xx$ and $\uu(j) = a$.

Lastly, for all $\epsilon > 0$, there exists $N(\xx, a)$ such that for all $N \geq N^j(\xx, a)$ and we can write $\left| \hat{q}_N^j\left(\xx, a \right) - \hat{q}^j_{\mathcal{S}_L}\left(\xx, a \right) \right| < \epsilon$.
Consequently, for $\epsilon >0$, we have $\left\| \hat{q}_N^j - \hat{q}^j_{\mathcal{S}_L} \right\|_{\infty} < \epsilon$.
for all $N \geq n(j) = \max_{\xx, a} N^j(\xx,a)$. $\qedproof$

\end{document}